\documentclass[a4paper,twocolumn,english,conference]{IEEEtran}
\pdfoutput=1
\usepackage[T1]{fontenc}
\usepackage{color}
\usepackage{amssymb}
\usepackage{graphicx}

\makeatletter

\providecommand{\tabularnewline}{\\}

\usepackage[caption=false,font=footnotesize]{subfig}
\usepackage[lined,boxed,commentsnumbered,ruled]{algorithm2e}
\usepackage{babel}

\@ifundefined{showcaptionsetup}{}{%
 \PassOptionsToPackage{caption=false}{subfig}}
\usepackage{subfig}
\makeatother

\usepackage{babel}
\begin{document}

\title{A Networked Swarm Model for UAV Deployment in the Assessment of Forest
Environments}

\author{\IEEEauthorblockN{Matthias R. Brust\IEEEauthorrefmark{1}\IEEEauthorrefmark{2}}\IEEEauthorblockA{\IEEEauthorrefmark{1}Computer Science Division\\
Technological Institute of Aeronautics, Brazil\\
Louisiana Tech University, Ruston, USA\\
Email: mbrust@latech.edu}\and \IEEEauthorblockN{Bogdan M. Strimbu\IEEEauthorrefmark{2}\IEEEauthorrefmark{3}}\IEEEauthorblockA{\IEEEauthorrefmark{2}School of Forestry\\
Louisiana Tech University, Ruston, USA\\
\IEEEauthorrefmark{3}National Forest Inventory, ICAS Bucharest, Romania\\
Email: strimbu@latech.edu}}
\maketitle
\begin{abstract}
Autonomous Unmanned Aerial Vehicles (UAVs) have gained popularity
due to their many potential application fields. Alongside sophisticated
sensors, UAVs can be equipped with communication adaptors aimed for
inter-UAV communication. Inter-communication of UAVs to form a UAV
swarm raises questions on how to manage its communication structure
and mobility. In this paper, we consider therefore the problem of
establishing an efficient swarm movement model and a network topology
between a collection of UAVs, which are specifically deployed for
the scenario of high-quality forest-mapping.

The forest environment with its highly heterogeneous distribution
of trees and obstacles represents an extreme challenge for a UAV swarm.
It requires the swarm to constantly avoid possible collisions with
trees, to change autonomously the trajectory, which can lead to disconnection
to the swarm, and to reconnect to the swarm after passing the obstacle,
while continue collecting environmental data that needs to be fused
and assessed efficiently.  

In this paper, we propose a novel solution to the formation flight
problem for UAV swarms. The proposed method provides an adaptive and
reliable network structure, which maintains swarm connectivity and
communicability. These characteristics are needed to achieve a detailed
and accurate description of the environment from the data acquired
by the UAV swarm. 

The main characteristics of our approach are high scalability regarding
the number of UAVs in the swarm and the adaptive network topology
within the swarm.\textcolor{red}{}

\textcolor{red}{}
\end{abstract}

\section{Introduction}

Unmanned Aerial Vehicles (UAVs) have gained popularity in many industries
due to their vast potential applications, such as survey, search and
rescue, agriculture, or forestry \cite{kendoul2012survey}. In the
last decade, a significant amount of research was carried on using
and processing the information supplied by a single UAV. However,
the reduced costs of commercial-grade UAV facilitates usage of interconnect
multiple UAVs in an adaptive and autonomous acting system \cite{burkle2011towards}. 

Such a UAV swarm is capable of accomplishing tasks which one UAV either
fulfills with difficulty, such as accurate determination of the location
for an object, or fails to accomplish altogether, such as mapping
of inaccessible caves or dense rain forest, assessment of real-time
environmental processes, or wildlife monitoring \cite{danoy2015connectivity}. Furthermore, compared
to a one UAV, a UAV swarm is able not only to solve more tasks, but
also to reduce the time of executing various activities and to increase
the quality of collected data. Additionally, if the task requires
navigational autonomy within an unknown or difficult environment,
a UAV swarm offers robustness through redundancy and self-organization,
which cannot be achieved by deploying one UAV \cite{danoy2015connectivity}. 

A particular case of a demanding flying environment is represented
by the interior of a forest. Data collection using sensors carried
above the forest canopy was executed for more than one century \cite{campbell2011introduction},
but accurate resource assessment still eludes foresters, as little
useful information can be obtained from sources located outside forest
\cite{shrestha2012estimating}. The limited success in describing
forest from afar is induced by the lack of algorithms that accurately
classify the information remotely sensed acquired in elementary components
(e.g., trees, shrubs, stem).

A UAV swarm can significantly increase the productivity and accuracy
of data describing the forest as a set of three dimensional objects,
each having multiple attributes attached (such as taper, infestation
with damaging biotic agents, or thermal radiance) \cite{strimbu2011analytical}.
An illustration of a deployment of a UAV swarm in a forest area is
shown in Fig. \ref{fig:Sensing,-collecting,-data}.  Each UAV maintains
communication capabilities with neighboring UAVs which are in its
transmission range. The UAVs adapt to the unknown environment, circumvent
obstacles, and map the forest with a variety of sensors. Afterwards,
the swarm leader UAV transmits the collected and aggregated data to
the base station or operator for data analysis and advanced post-processing.\textcolor{red}{}

However, the advantages promised by an autonomous UAV swarm face challenges
that come with the efficient swarm formation and communication preservation.
The set of UAVs should exhibit a swarm-like behavior that provides
an adaptive and reliable network structure while fulfilling the required
tasks for assessment of environment (e.g., tree dimensions, amount
of light, or spatial variation of humidity inside the forest) \cite{brust2012clustering,brust2011multi}.

In this paper, we propose a solution for establishing swarm behavior
and a network topology between a collection of UAVs, which are specifically
deployed for the scenario of high-quality forest-mapping. Our approach
applies a leader election algorithm to a set of autonomous micro UAVs.
The leader will have tasks such as gathering information from the
swarm collective and leading the swarm to the destination. The leader
UAV is additionally in charge to communicate to the base station (cf.
Fig. \ref{fig:Sensing,-collecting,-data}). The UAVs in the formation
sense and observe the events in the environment and the more powerful
leader collects and processes information from them and reacts in
case of obstacle avoidance and acts as it the cause for route planning
and maneuvers. 

\textcolor{red}{}

\section{Related Work\label{sec:Related-Work}}

Although there is high interest in applications for UAVs in environmental
monitoring, such as forest-mapping, airborne surveillance or space
exploration, most approaches are focused on the usage of one UAV.
One of the reasons is that formation, coordination and control of
a UAV swarm system containing multiple UAVs induces additional challenges,
such as three dimensional movements, which can be $NP$-hard.

Existing approaches for optimizing formation acquisition and maintenance
mostly focus on\textit{ coverage problems} \cite{Brust2009a,schleich2013uav}.
A variant of the coverage problem is the \emph{maximal coverage problem},
where packing of a maximum number of circles is required. For the
two dimensional case the problem has a polynomial time solution \cite{huang2005coverage}.

When considering three dimensions, the question for optimal node positioning
is called the \textit{sphere packing problem} \cite{hales1992sphere}.
A related problem in geometry is the \emph{kissing number problem
}\cite{musin2006kissing}, which is the number of non-overlapping
unit spheres arranged such that each sphere touches another.

The properties of network topologies resulting from random deployment
of nodes in a three-dimensional area are studied by Ravelomanana \cite{ravelomanana2004extremal}.
Ravelomanana considers the \textit{k-connectivity problem}, which
looks for the lower bound of the transmission range $r$, so that
every node has at least $k$ direct neighbors.

Schleich et al. \cite{schleich2013uav} propose a decentralized and
localized approach for UAV mobility control, which optimizes the network
connectivity. The approach maintains the connectivity via a tree-based
overlay network, whereby the root is the base station. Their empirical
results show that the maintenance of the connectivity can have a negative
impact on the coverage while the overall connectivity can improve.

The main feature of our approach is that a multi-path communication
structure is maintained while using only local information, thus its
preserving locality. Consequently, the formation can effectively span
over a significantly wider area. Furthermore, we also introduce a
leader navigator in the swarm movement, which distinguishes our approach
from classical swarm modeling approaches.

\begin{figure}
\begin{centering}
\includegraphics[width=0.9\columnwidth]{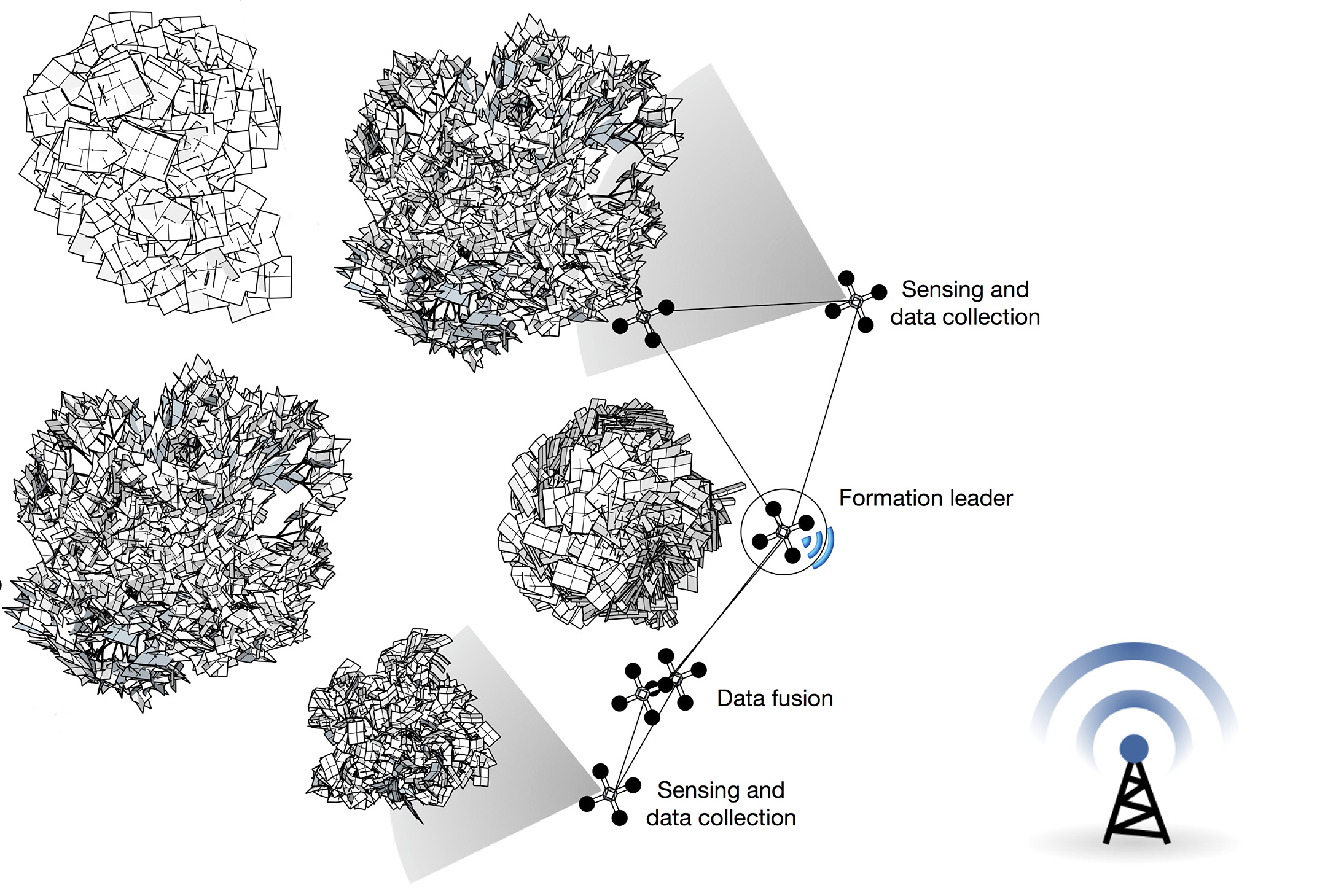}
\par\end{centering}
\caption{A UAV swarm is deployed in a forest environment to collect and fuse
data for transmission to ground controller.\label{fig:Sensing,-collecting,-data}}
\end{figure}
The traditional remote sensing techniques employed in forest resources
assessment and monitoring rely on imaginary data often operating in
the visible spectrum, which fail to provide useful information on
cloudy days. Although remedies have been found, such as combining
visible and infrared light \cite{hansen2008method} a more direct
approach is the usage of UAVs, specifically when flying under the
canopy. A significant reduction in data acquisition costs while increasing
accuracy occurs when instead of one flying entity multiple UAVs are
used \cite{frey2008physicomimetics,sanchez2014system}, which conducted
studies on land use in the Congo Basin, where the ground was often
obscured by clouds and, therefore, developed ways to create composite
pictures of sources with visible and infrared light. 

While UAV swarms approaches are popular in military, communication,
and marine applications \cite{wei2013operation,hauert2009evolved},
only few applications are focused on forestry, mainly in forest fire
surveillance \cite{casbeer2006cooperative}. The scenario considered
throughout this paper aims forest-mapping of healthy trees, and introduces
a set of particular challenges to the UAV swarm movement, such as
a continuous change of the communication topology due to the high
occurrence of trees in a relatively reduced area, which differentiates
it to existing approaches.

\section{Swarm System Model\label{sec:Swarm-System-Model}}

In this section, we define a basic swarm system model and the formal
notions used to describe it. 

\subsection{UAV Swarm System Model}

\begin{figure}
\begin{centering}
\includegraphics[width=0.95\columnwidth]{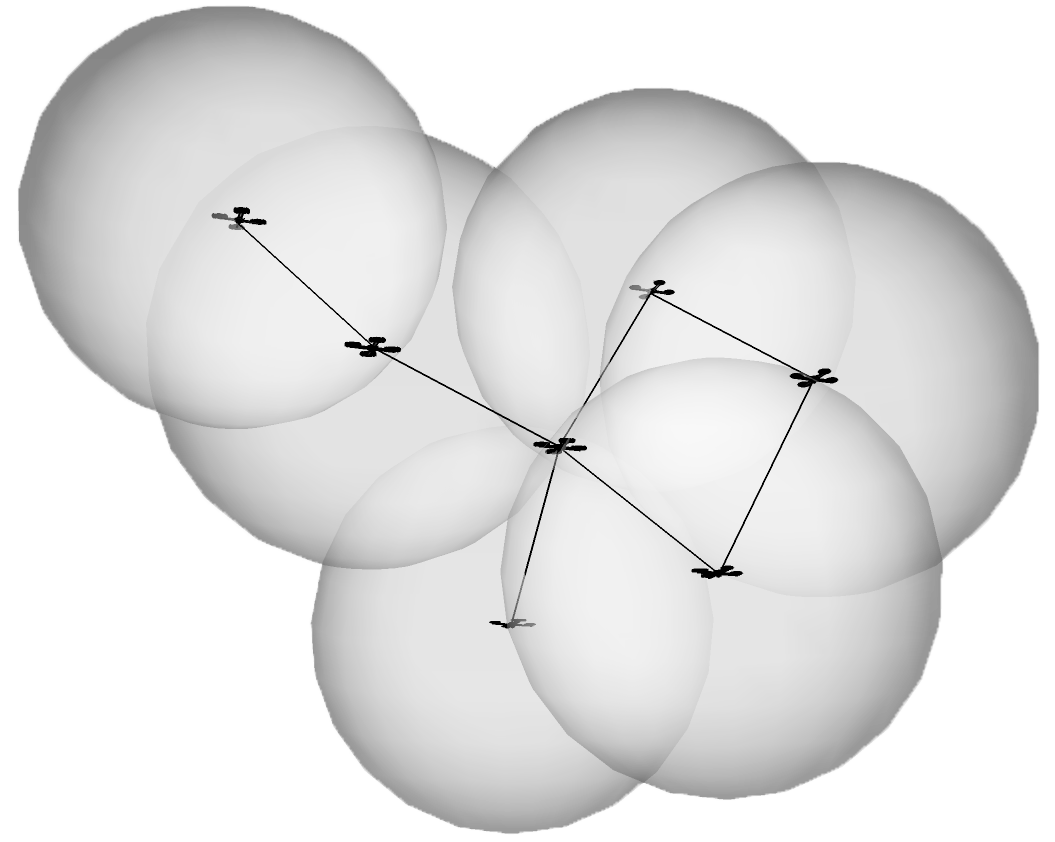}
\par\end{centering}
\caption{A networked UAV swarm. The transmission range is illustrated by the
shaded spheres around each UAV. UAVs in mutual transmission range
are able to communicate and transfer collected data (continuous line).
\label{fig:Spheres-illustrate-the}}
\end{figure}

The communication network of a UAV swarm $S$ is represented by a
symmetric Euclidean graph $G=\left(V,E\right)$ constructed such that
$V\in{\mathbb{R}}^{{\rm 3}}$ is a set of UAV nodes in a three dimensional
bounded region with side length $l$. The UAV nodes are deployed according
to a deployment model $D$. The links in set $E$ of the graph $G$
fulfill the condition that for any pair $u,v\in V$ of nodes, $dist\left(u{\rm ,}v\right)\le r\Longrightarrow\left\{ u,v\right\} \in E$
and $dist\left(u{\rm ,}v\right){\rm >}r\Longrightarrow\left\{ u,v\right\} \notin E$,
where $r$ is the effective transmission range for each $v\in V$.
Each UAV $v\in V$ can relocate to any position in $V\in{\mathbb{R}}^{{\rm 3}}$.
An exemplary swarm network topology, which results from our swarm
system model is shown in Fig. \ref{fig:Spheres-illustrate-the}. 

Our work addresses the connections only and does not measure energy
efficiency from a signal-point of view, but from a topological point
of view; wherefore, we do not describe a radio propagation model here.

For each UAV node $v\in V$ let there be a neighboring list $Neigh(v)\subset V$,
which is the set of UAVs directly connected to UAV $v$, such that
$\forall u\in Neigh(v),d(v,u)\leq r$. The neighboring list $Neigh(v)$
is created initially and is updated with frequency $f$, since the
neighborhood $Neigh(v)$ of a node $v$ is subject to change.We further
assume that each UAV node $v\in V$ can communicate exclusively with
its current direct neighbors $Neigh(v)$ (\textit{1-hop neighbors}). 

Since only localized communication is used, there is no need to establish
multi-hop control communication. This specification guarantees complexity
efficiency in our locality-preservation approach.

The swarm leader uses different ways to deliver data, which can be
the submission of data if satellite or cellular communication is available
at the base station. For this, the leader can stop periodically the
swarm activity to fly above the tree crowns for data submission. The
data can also be delivered after returning to the spatial proximity
to the base station. However, in this paper we exclusively deal with the swarm behavior
and networking aspects within the swarm. Swarm-to-base station communication
is subject of future work.

One crucial objective for the deployment of a UAV swarm is to obtain
accurate location information of the UAVs and the objects under investigations
such as trees, the trees' crown and canopy.

However, the inherent density of a forrest imposes an extreme challenge
on on-site data collection: GPS is not a viable option in terms of
availability and efficiency for UAVs. Additionally, thick forest structure
may block satellite communication from time to time as the UAVs fly
below the crowns and canopy.

Therefore, an autonomous UAV swarm needs to rely on a relative positioning
system, where only a small subset of the UAVs relies on GPS, for planning
or maneuvering. 

Therefore, our model assumes that for each swarm $S_{i}$ only the
leader UAV $S_{i}\left(L\right)$ is in charge of absolute positions
(e.g. GPS data or preinstalled maps). Therefore, the swarm leaders
have \textit{a priori} knowledge of their positions.

\section{Approach: A Networked Swarm Model\label{sec:Proposed-Approach}}

Various requirements and challenges exist to assess unpredictable
environments such as a forest using a UAV swarm. Differentiating features
such as autonomous control or the continuous motion during the flight
require the swarm to have a high resistance threshold concerning changes
in positions, and the swarm must reorganize obstacles and overcome
them. The algorithm for describing swarm movement and maintaining
the network topology between the UAVs in the swarm is localized and
fully distributed. Additionally, our approach does not require communication
with a base station or the global positioning system.  

\subsection{Leader UAV and swarm communication network}

The UAV swarm formation process needs to consider that in restricted
settings, the leader UAV will be predetermined due to higher-energy
resources and/or communication capabilities. Moreover, we aim at providing
a scalable approach, in which the leader election is triggered in
the initialization phase and re-triggered periodically during the
flight. The possibility of changing the leader UAV allows the swarm
to readjust to environmental conditions, and to maximize its operation
efficiency.

In the first part of Algorithm \ref{algorithm1}, we deal with the
communication network in the UAV swarm. The leader UAV and, subsequently,
the communication structure in the swarm is determined. Initially,
weights are assigned to each UAV, which change according to the conditions
in Algorithm \ref{algorithm1} to determine its function in the swarm
communication network. The described procedure explicitly supports
changes of UAVs in the network position, therefore, providing an adaptive
and self-organizing approach.

\subsection{Swarm movement and formation control}

Our approach requires knowledge of destination only by the leader
UAV. The additional swarm members follow the leader in a collision-free
manner by adjusting their velocity according to the conditions that
describe the behavior of the swarm.

To achieve this leader-follower model, we significantly altered the
leaderless\emph{ Boid model} \cite{reynolds1987flocks}, and build
a distributed formation-featured model that fulfills the swarm requirements
for our application scenario. We propose the following extended conditions
for the UAV swarm with a leader:
\begin{enumerate}
\item The leader UAV receives its current position and destination.
\item Each UAV (including the leader UAV) aligns with its direct neighbors.
\item Each UAV (including the leader UAV) maintains direct connectivity
with its neighbors.
\item Each UAV (including the leader UAV) avoids collision with any UAV.
\item The leader UAV  approaches the destination.
\end{enumerate}
\begin{algorithm}
\LinesNumbered 
\SetKwInOut{Input}{Input}
\SetKwInOut{Output}{Output} 
\Input{A set of UAVs $U$ with positions Pos and velocities Vel,
$weightLimit$, $vel_{leader}$, $leaderID$, $d_{1}$, $d_{2}$} 
\Output{A set of UAVs $U$ with adjusted positions and velocities} 
\BlankLine 
\ForEach{UAV $u\in U$}{ 
   $u_{w}$ $\leftarrow$ Weight($u$)\; 
   $N(u)$ $\leftarrow$ FindDirectNeighbors($u$)\; 
   \tcc{Communication network} 
   R1 $\leftarrow$ Max$(\{\forall u\in N(u):\mbox{Weight}(u)\})$\; 
   R2 $\leftarrow$ Min($\{\forall u\in N(u):\mbox{Weight}(u)\})$\; 
   R3 $\leftarrow n\in N(u)$ if $\mbox{Weight}(n)=leaderID$\;
   \If{$R2<u_{w}$}{
      $u_{w}\leftarrow R2+1$\; 
      \If{$R1=weightLimit$}{
         $u_{w}\leftarrow leaderID$\; 
         \If{$R2\geq u_{w}\land u_{w}\neq leaderID$}{
            $u_{w}\leftarrow R2+1$\; 
            \If{$u_{w}=leaderID\land R3\notin\varnothing$}{ 
               $u_{w}\leftarrow u_{w}+1$\; 
            } 
         } 
      } 
   } 
   \tcc{Swarm behavior} 
   newVel() = $\left\{ \right\} $\;  
   $centreOfMass$ $\leftarrow$ $\sum\nolimits _{n_{i}\in N(u)}$Pos$(n_{i})$\; 
   newVel(1) $\leftarrow$ $(centreOfMass-$Position$(u))*0.2$\;
   \If{There is a neighbor $n\in N(u)$ with $|\mbox{Pos}(n)-\mbox{Pos}(u)|<100$}{ 
      newVel(2) $\leftarrow$ newVel($2$) - (Pos($n$) - Pos($u$))\; 
   }  
   newVel(3) $\leftarrow$ $\sum\nolimits _{n\in N(u)}$Vel$(n)\frac{1}{|N|-1}$\; 
   newVel(3) $\leftarrow$ $($newVel$(3)-$Vel$(u))*0.2$\;  
   \If{$\mbox{Weight}(u)=leaderID$}{
      $diff$ $\leftarrow$ $d_{2}$ - Pos($u$)\; 
      \eIf{ $diff<5$}{
         newVel(4) $\leftarrow$ $\left\{ 0,0,0\right\} $\; 
      }{ 
         newVel(4) $\leftarrow$ $diff/$Norm$[diff]*vel_{leader}$\; 
      } 
   } 
   Vel($u$) $\leftarrow$ Vel($u$) + $\sum\nolimits _{v\in newVel}v$\; 
   Pos($u$) $\leftarrow$ Move(Pos($u$),Vel($u$))\;
} 

\caption{A networked swarm model}\label{algorithm1}
\end{algorithm}

These five conditions represent a complete description of the UAV
movement. The first condition ensures that coordinates of the current
position and the destination are known by the designated leader UAV.
 The leader respects all conditions, while moving towards destination.
The second condition describes the behavior of a UAV in relationship
with its neighbors and is responsible for alignment maintenance of
the local group movement behavior. This condition is important to
spatially organize the swarm and, in particular, to offer an enhanced
coverage of the object under investigation (e.g. tree). Conditions
three and four describe two opposing behaviors; the former reinforces
UAV movements that preserve the swarm (i.e. relationships to its momentarily
direct neighbors), while the latter ensures that swarm\textquoteright s
members are at a collide avoidance distance (i.e. preset minimum distance).
The fifth condition serves primarily to direct the swarm to its destination.
The procedure implementing the five conditions is formally described
in the second part of Algorithm \ref{algorithm1}.

Local rules or conditions limit the increase in complexity induced
by scenarios with large numbers of UAVs. Therefore, the proposed approach
is scalable, as it can be applied to swarms with either few or multiple
UAVs.

\section{Simulations and Results\label{sec:Simulations-and-Results}}

\paragraph*{Settings\label{par:Settings}}

We refer to the notions of the basic swarm system model in \ref{sec:Swarm-System-Model}
for the descriptions of the simulation settings.

We define the UAV swarm $S$ as a graph $G=\left(V,E\right)$ constructed
such that $V\in{\mathbb{R}}^{{\rm 3}}$ is a set of UAV nodes in a
three dimensional bounded region with side length $l=200\,m$. The
number of UAVs used is $|V|=8$ and the transmission range $r=55\,m$.
The set $E$ is determined by the condition described in \ref{sec:Swarm-System-Model},
which states that for any pair $u,v\in V$ of nodes $dist\left(u{\rm ,}v\right)\le r\Longrightarrow\left\{ u,v\right\} \in E$
and $dist\left(u{\rm ,}v\right){\rm >}r\Longrightarrow\left\{ u,v\right\} \notin E$.

The UAV nodes are positioned according to the deployment model $D$
with coordinates $d_{1}=\{25,25,25\}$ (i.e., deployment) and $d_{2}=(100,100,100)$
(i.e., destination) whereby $d_{1},d_{2}\in\mathbb{R}^{{\rm 3}}$.
The leader UAV velocity is $vel_{leader}=0.7\,\frac{m}{s}$.

The neighboring list $Neigh(v)\subset V$ will be created initially
so that $\forall u\in Neigh(v),d(v,u)\leq r$ and updated with frequency
$f=10s^{-1}$. 

The UAVs start in the simulation from virtually the same position.
The leader UAV moves from the deployment coordinates $d_{1}$ to the
destination coordinates $d_{2}$ with velocity $vel_{leader}$. Algorithm
\ref{algorithm1} (re-)calculates the velocity vectors of each UAV
such that the swarm maintains connectivity, but avoids collisions.

\begin{figure*}
\begin{centering}
\subfloat[Initial deployment and leader election\label{fig:Initial-deployment}]{\begin{centering}
\includegraphics[width=6.6cm]{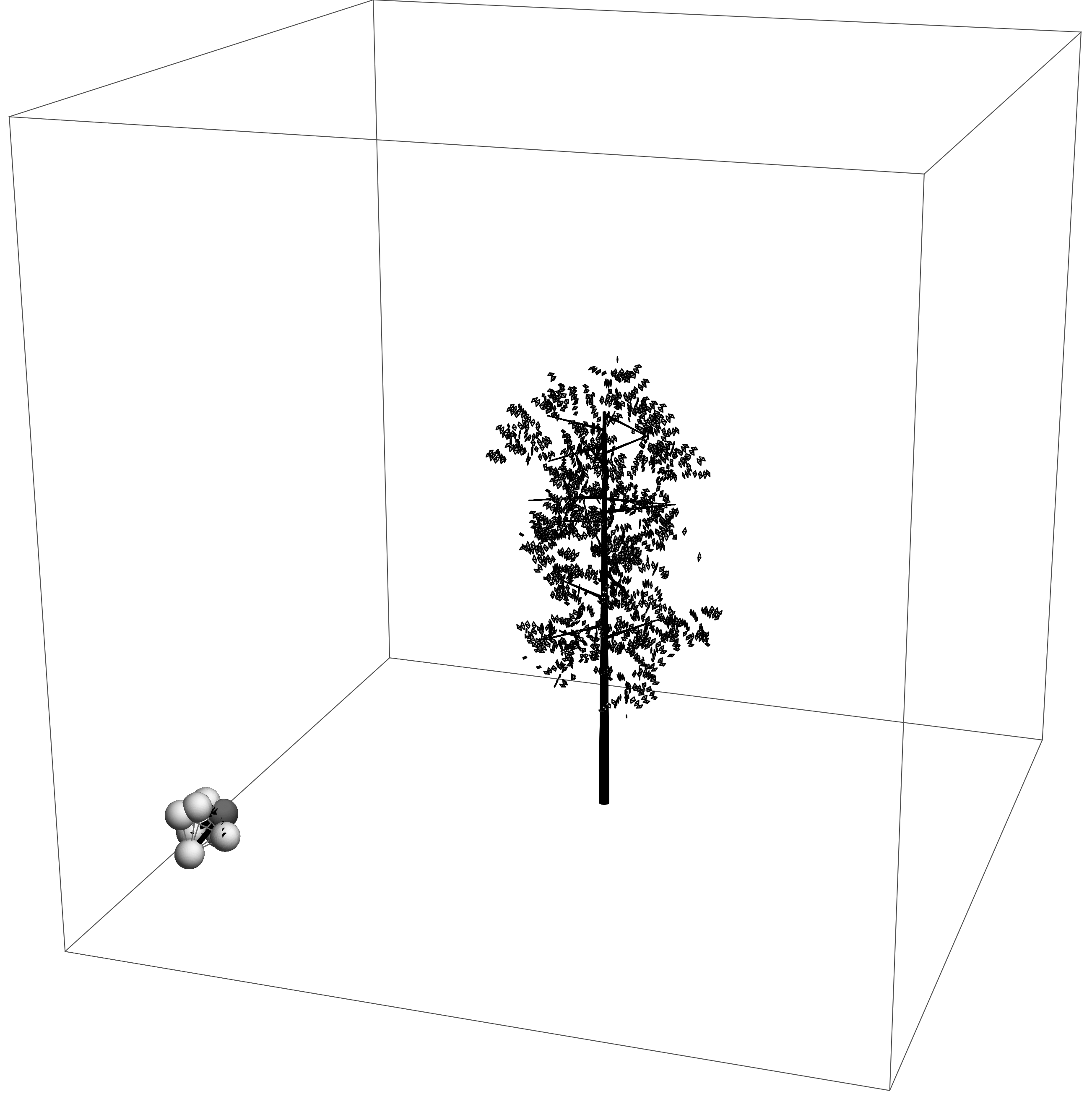}
\par\end{centering}
}~~~~~~~~~~~~~\subfloat[Establishing communication network and navigation\label{fig:Leader-election-and}]{\begin{centering}
\includegraphics[width=6.6cm]{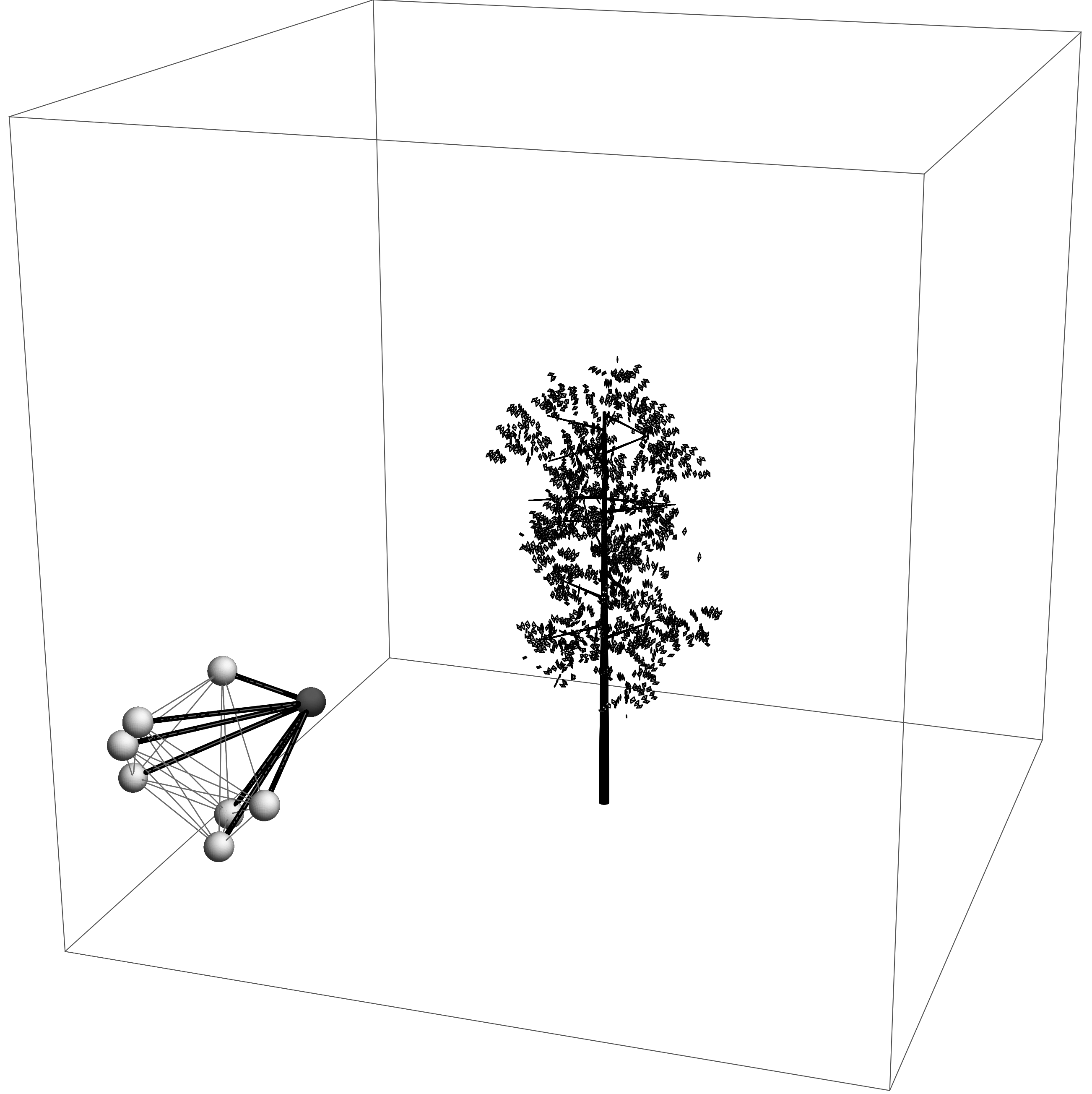}
\par\end{centering}
}
\par\end{centering}
\begin{centering}
\subfloat[Navigation and maintenance of communication\label{fig:Navigation-and-maintenance}]{\begin{centering}
\includegraphics[width=6.6cm]{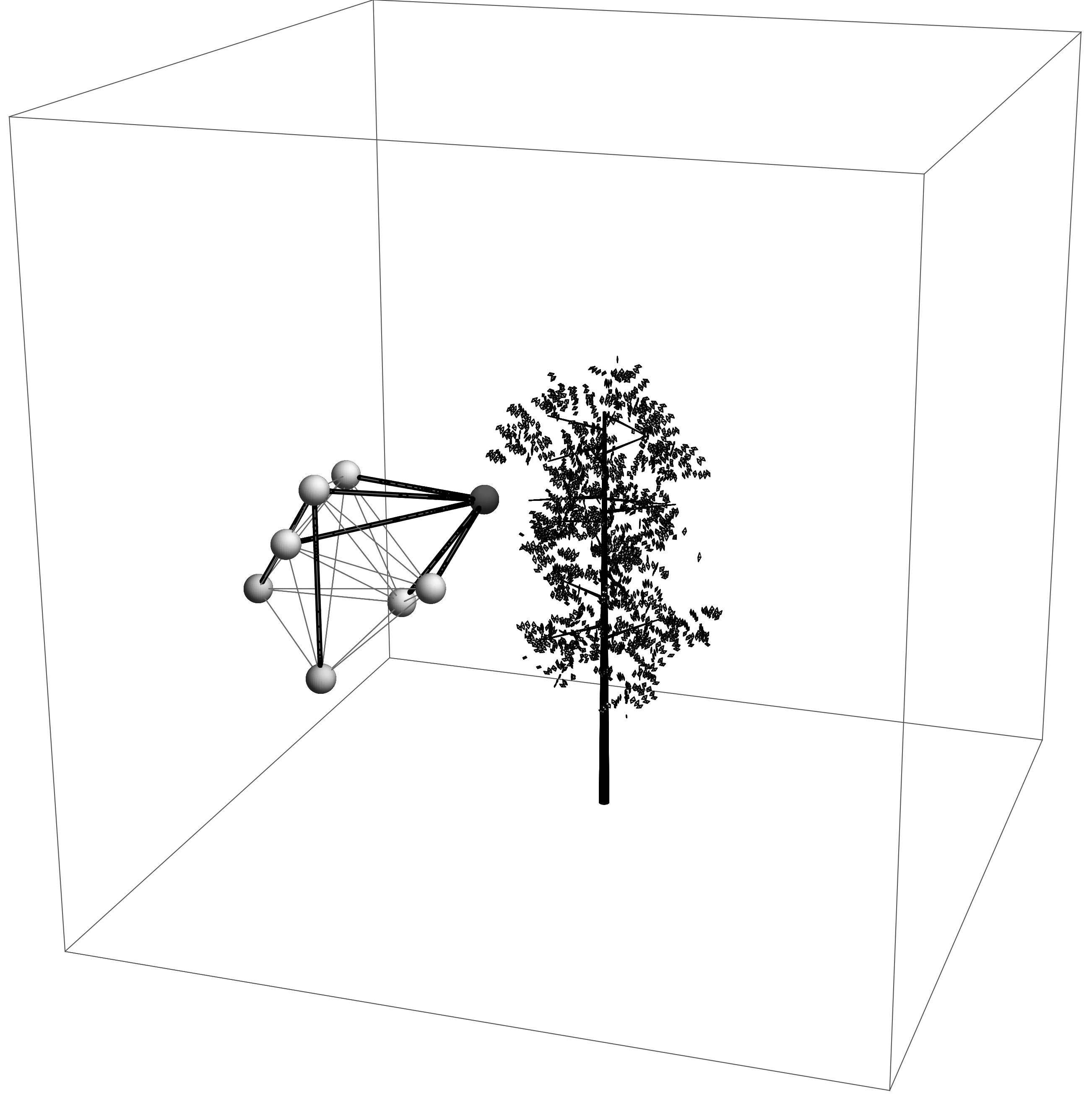}
\par\end{centering}
}~~~~~~~~~~~~~\subfloat[Positioning of the UAV swarm\label{fig:Positioning-of-the}]{\begin{centering}
\includegraphics[width=6.6cm]{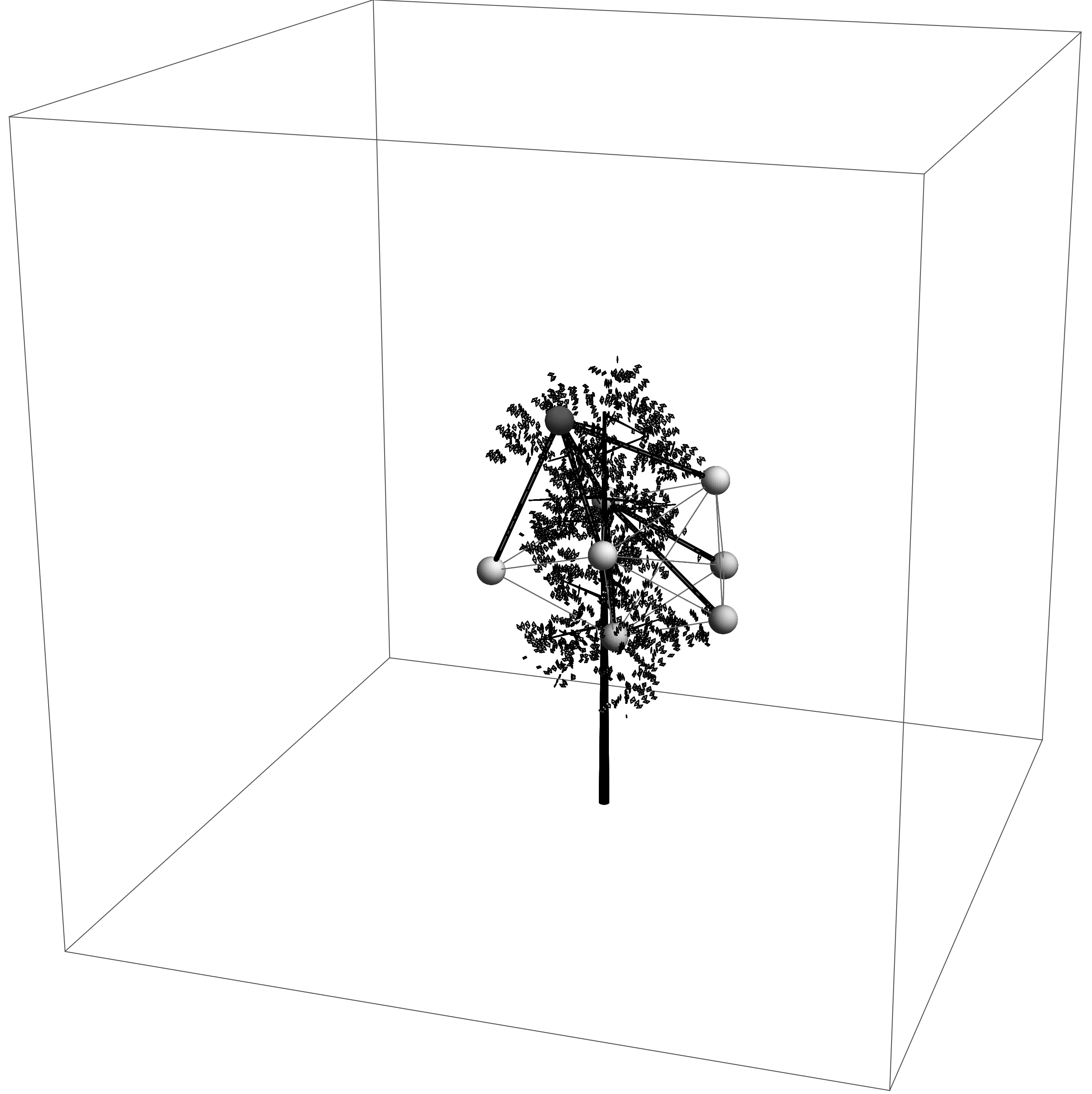}
\par\end{centering}
}
\par\end{centering}
\caption{Simulation of a networked UAV swarm with self-organizing communication
network as described in Algorithm \ref{algorithm1}. The leader UAV
is dark colored and the established communication network is shown
in bold continuous lines.\label{fig:Simulation-results-of}}
\end{figure*}

\paragraph*{Simulation behavior}

We implemented a UAV swarm simulator to visually and empirically verify
the usability and performance of our proposed approach described in
Algorithm \ref{algorithm1}. Simulation results for the simulation
settings described in Section \ref{par:Settings} are shown in Fig.
\ref{fig:Simulation-results-of}.

Figure \ref{fig:Initial-deployment} shows that the UAVs are positioned
randomly around the deployment coordinates $d_{1}=\{25,25,25\}$.
Noteworthy, the algorithm supports random position for each UAV in
the simulation area.

Figure \ref{fig:Leader-election-and} shows that the leader election
has been executed and that the leader UAV is directly connected to
each of the UAVs in the swarm. Additionally, the leader UAV starts
moving towards the destination coordinates $d_{2}=\{100,100,100\}$. 

Figure \ref{fig:Navigation-and-maintenance} shows that the leader
UAV pushes the swarm towards the destination $d_{2}$. Observe that
only the leader knows the destination coordinates and the rest of
the swarm follows the leader, while maintaining the swarm behavior.

For initiation of the coverage procedure, we assume that the swarm's
movement to the destination has been realized with a lower transmission
range, which keeps the UAVs closer and the swarm compact. Arriving
at the destination coordinates $d_{2}$ the UAVs transmission is extended
to its full range. Figure \ref{fig:Positioning-of-the}  shows the
leader UAV arriving at the tree, which represents the destination.
Subsequently, the UAV swarm forms a coverage around the tree, while
maintaining full connectivity to the entire swarm. Each UAV is connected
directly or over several hops to the leader UAV.

\paragraph*{Time assessment}

Although, the leader UAV pushes the UAV swarm towards the destination
$d_{2}$, the swarm movement procedure in Algorithm \ref{algorithm1}
induces a delay in the collective's movement to avoid disconnection
but at the same time to avoid collision between UAVs. Therefore, we
are interested in the movement delay caused by our approach compared
to a straight flight of solely the leader UAV from $d_{1}$ to $d_{2}$. 

For this, we conducted five runs of the simulation, each with a different
number of UAVs ($n=4,n=8,n=12$). We measured the time needed for
the swarm to arrive at the tree ($d_{2}$) and compared this to the
theoretical time, which is the time needed by only one UAV to fly
directly with velocity $vel_{leader}$ from $d_{1}$ to $d_{2}$.
This quantitative measure reflects applicability and scalability of
the proposed approach, since the swarm conditions impose influence
on the overall task.

\begin{table}
\begin{centering}
\begin{tabular}{|c|c|}
\hline 
\textbf{Parameter} & \textbf{Value}\tabularnewline
\hline 
\hline 
Euclidean distance to destination & 129.9\tabularnewline
\hline 
Velocity vector of the leader UAV & 0.7\tabularnewline
\hline 
Theoretical time to destination & 185.57\tabularnewline
\hline 
\end{tabular}
\par\end{centering}
\smallskip{}

\begin{centering}
\begin{tabular}{|c|c|c|c|}
\hline 
\textbf{Run} & \textbf{$n=4$} & \textbf{$n=8$} & \textbf{$n=12$}\tabularnewline
\hline 
\hline 
1  & 190 & 190 & 263\tabularnewline
\hline 
2 & 195 & 199 & 283\tabularnewline
\hline 
3 & 192 & 199 & 225\tabularnewline
\hline 
4 & 196 & 194 & 234\tabularnewline
\hline 
5 & 193 & 201 & 218\tabularnewline
\hline 
Average & 193.2 & 196.6 & 244.6\tabularnewline
\hline 
\end{tabular}
\par\end{centering}
\smallskip{}

\caption{Temporal performance of the UAV swarm.\label{tab:Temporal-performance-of}}
\end{table}

\begin{figure}[h]
\begin{centering}
\includegraphics[width=8.5cm,height=6.2cm]{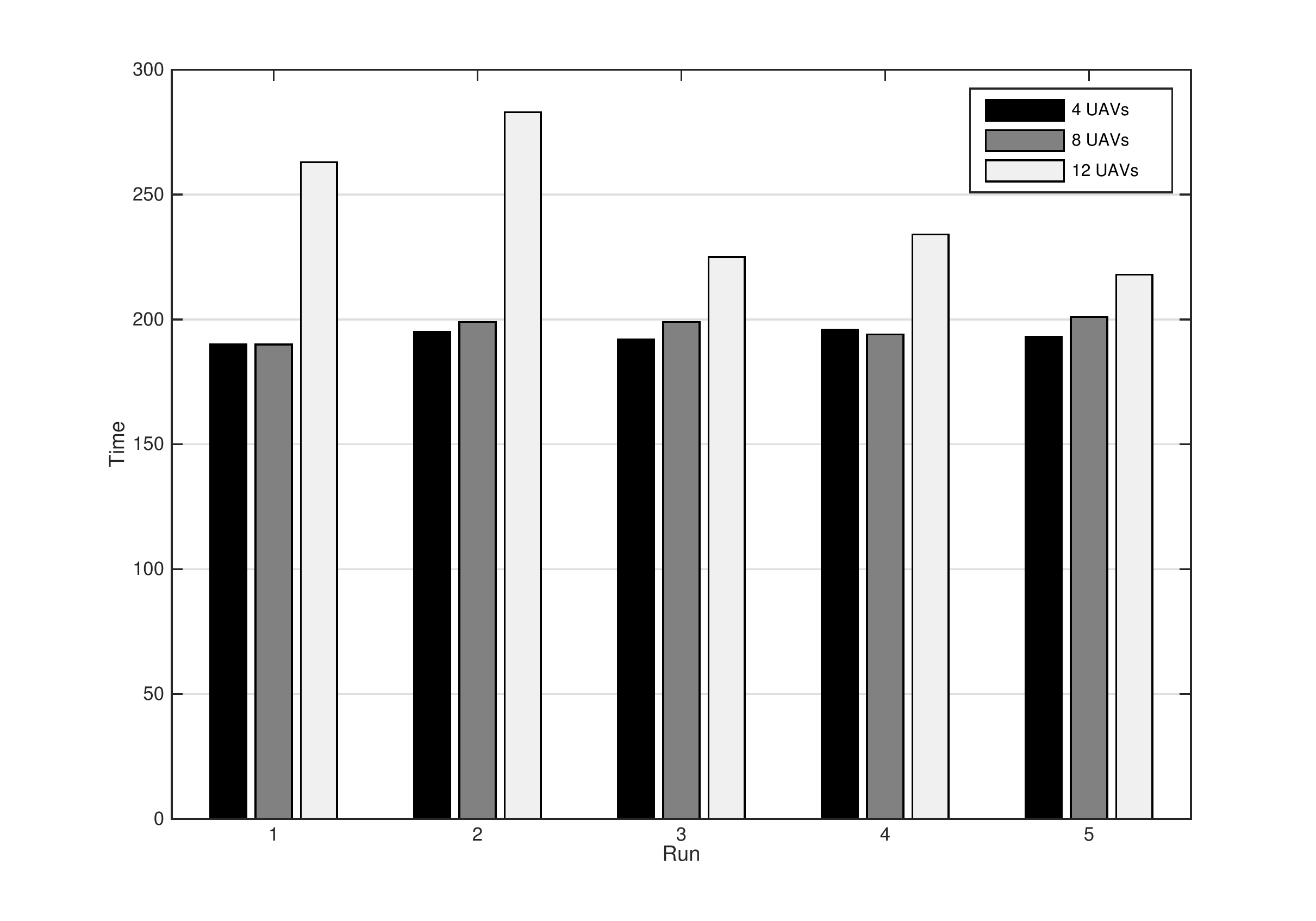}
\par\end{centering}
\caption{Presentation of the UAV swarm's temporal performance.\label{fig:Graphical-presentation-of}}
\end{figure}

Results in Table \ref{tab:Temporal-performance-of} and Fig. \ref{fig:Graphical-presentation-of}
show that the flying time from $d_{1}$ to $d_{2}$ (i.e. simulation
time in discrete steps) does not differ significantly when 4 or 8
UAVs are deployed (i.e., $1.8\%$ from direct flight time), which
is $4.1\%$ time increase for $n=4$ and $5.9\%$ for $n=8$. However,
for the case $n=12$, we observe a considerable increase in time (i.e.,
31\%). For the envisioned scenario (i.e., the forest environment where
we focus on assessing one or several trees) the number of UAVs below
$10$ is realistic and practical.  

\section{Conclusions\label{sec:Conclusions}}

We propose an approach for the formation flight problem for UAV swarms,
which maintain connectivity and communicability. The two characteristics
are important for detailed, precise, and accurate data collection
and aggregation, and play a crucial role in assessing the forest environment.
Our approach is based on local decision-making, and, therefore, assures
low message and computation complexity, while providing scalability
for a large number of UAVs in the swarm. Results indicate that the
proposed swarm movement is almost as fast to reach the destination
location as the theoretical approach comprising of just one UAV.

However, melding the aspects of swarm movement and communication network
is more than a matter of concatenating two isolated aspects. Additional
issues of optimal swarm positioning, such as ad hoc data fusion and
object assessment (e.g. localization and mapping) will be addressed
in future work.

\bibliographystyle{ieeetr}
\bibliography{Swarm124}

\end{document}